\pdfoutput=1
\documentclass[11pt]{article}

\usepackage{authblk}

\usepackage{EMNLP2022}

\usepackage{times}
\usepackage{latexsym}
\usepackage{natbib}
\usepackage{CJKutf8}
\usepackage{graphicx}
\usepackage{array}
\usepackage{svg}
\usepackage{multirow, booktabs}
\usepackage{amsmath,amssymb}
\usepackage{pgfplots}
\usetikzlibrary{patterns}
\usepackage{siunitx}

\newcommand\blfootnote[1]{%
  \begingroup
  \renewcommand\thefootnote{}\footnote{#1}%
  \addtocounter{footnote}{-1}%
  \endgroup
}

\usepackage[T1]{fontenc}
\usepackage[utf8]{inputenc}
\usepackage{fdsymbol}
\usepackage{microtype}

\usepackage{inconsolata}

\title{ Professional Presentation and Projected Power: \\A Case Study of Implicit Gender Information in English CVs }

\author[1*]{\bf{Jinrui Yang}}
\author[2*$\varheartsuit$]{\bf{Sheilla Njoto}}
\author[1]{\bf{Marc Cheong}}
\author[2]{\bf{Leah Ruppanner}}
\author[1]{\bf{Lea Frermann}}
\affil[1]{School of Computing and Information Systems, University of Melbourne}
\affil[2]{School of Social and Political Sciences, University of Melbourne}
\affil[ ]{\texttt{\{jinruiy, snjoto\}@student.unimelb.edu.au}}
\affil[ ]{\texttt{\{marc.cheong, leah.ruppanner, lea.frermann\}@unimelb.edu.au}}

\begin{document}
\maketitle
\begin{abstract}
Gender discrimination in hiring is a pertinent and persistent bias in society, and a common motivating example for exploring bias in NLP. However, the manifestation of gendered language in application materials has received limited attention. This paper investigates the framing of skills and background in CVs of self-identified men and women. We introduce a data set of 1.8K authentic, English-language, CVs from the US, covering 16 occupations, allowing us to partially control for the confound occupation-specific gender base rates. We find that (1)~women use more verbs evoking impressions of low power; and (2)~classifiers capture gender signal even after data balancing and removal of pronouns and named entities, and this holds for both transformer-based and linear classifiers.\blfootnote{* Equal contribution}\blfootnote{$^\varheartsuit$ Corresponding author}
\end{abstract}

\newcommand{\sheilla}[1]{\textcolor{red}{\it Sheilla: #1}}
\newcommand{\jinrui}[1]{\textcolor{blue}{\it Jinrui: #1}}
\newcommand{\todomc}[1]{\textcolor{orange}{\it MC TODO: #1}}

\section{Introduction}

% intro paragraph
In this paper, we study word choice and implied power and agency in curriculum vitae (CVs) authored by men and women, combining lines of research that emerged from 
% paragraph on gendered word choice (1) causes; (2) effects/perception
a long research tradition in both the social sciences and, more recently, natural language processing %has proved a wealth of evidence for and characteristics of the differences 
~\cite{carli1990gender,lakoff1975linguistic,glick2018ambivalent}. %These differences manifest for instance in word choice. 
From a sociology perspective, it has been suggested that choices of words are influenced by the social status of the respective genders at a given moment in society~\cite{talbot2019language}. Women are known to use more communal forms of words and emotional connotations than men~\cite{brownlow2003gender,leaper2007meta,newman2008gender}, and that such choices reflect the different levels of power and influence both politically and economically~\cite{talbot2019language,leaper2007meta}. Conversely, the choice of language impacts how the reader {\it perceives} the entity described in the text. In particular, the choice of verbs has been suggested as an indicator of the perceived levels of {\it power} and {\it agency} of the described entity~\cite{sap-etal-2017-connotation}. 

Organisational scholars have long documented gender discrimination in employment~\cite{booth2010employers,heilman2012gender,steinpreis1999impact}. Sociological studies have repeatedly shown that women are evaluated more harshly than men especially in recruitment~\cite{moss2012science,NBER_discrimination,riach2006experimental}. Men tend to be assessed for their competence, while women are assessed based on characteristics (`likeability'), even when they demonstrate the same levels of qualifications, experience and education~\cite{rudman1998self,phelan2008}. \citet{Gaucher2011-rr} studied the impact of ``gendered wording'' in job advertisements on gender inequality in traditionally male-dominated occupations via content analysis, while~\citet{de2019bias} showed how gender signal in online  biographies lead to disparate performance in the task of occupation classification. Experience has shown that leaving hiring decisions to supposedly objective algorithms did not remove bias from the process -- both in real-world applications like Amazon's gender-biased automatic hiring tool~\cite{bogen2019all}, as well as a surge in research on showing and alleviating bias in NLP models~\cite{sun-etal-2019-mitigating}.
% To mitigate gender biases with human hirers, the determining indicator of gender (i.e., the applicant’s name), is scrubbed (blind hiring)~\cite{Manikandan}.  However, gender indicators become even more complex when automation is involved. 

% our data set
We present a data set of 1.8K %naturalistic,
human-written CVs and study differences in word choice and framing between men and women, and the extent to which classifiers are susceptible to gendered language. 
% We collected our data set from crowd-workers on Prolific Academic\footnote{\url{https://www.prolific.co/}}, a crowdsourcing platform which emphasizes fair treatment of workers and better quality control of data. 
Unlike prior studies which were either occupation-specific~\cite{parasurama-sedoc-2022-gendered} or used proxy data like online biographies~\cite{de2019bias}, we inspect application materials directly and cover 16 occupations  (Appendix~\ref{sec:app:occupations}) which allows us to study gender differences while partially controlling for the confound of %occupation-specific information due to the fact that men and women are unequally represented in most occupations.
occupation-specific base rates. However, we find that even within occupations, confounds remain as women tend to occupy lower ranking positions and men and women cluster in different types of fine-trained jobs within an occupation category~(Section~\ref{sec:tfidf_qual}). 

% how we treat gender: (1) self-identified; (2) binary
Our CV authors provide self-identified gender as part of our screening questions.\footnote{Participants chose from: [Man, Woman, Other/non-binary, prefer not to say].} Due to the very low number of "Other/non-binary" responses ($0.01\%$), we here only consider people self-identifying as male or female. We acknowledge that treating gender as a binary phenomenon is an oversimplification~\cite{Guo2020-hw}, but stress that our methodology extends to more inclusive notions of gender, and hope that our study inspires future work in this direction.

After presenting our data set (Section~\ref{sec:data}), we investigate gender signals in CVs in terms of overall word choice (Section~\ref{sec:tfidf_qual}); implied associations with power and agency (Section~\ref{ssec:powerandagency}); and predictive models' sensitivity to gender when trained on data from which gender-indicative signals were removed to different extents~(Section~\ref{sec:prediction}).

%%%%%%%%% uncomment this if at all possible
% In the following, we present our data set, before we leverage our cross-occupation CV data set to study the difference in power- and agency suggestive wording, and the ability of classifiers to predict gender from CV content. We conclude with a discussion of results and impact for future work in NLP and sociology.

\section{Dataset collection}
\label{sec:data}
% Our data collection process was approved by the local ethics review board (approval number 22062).\footnote{redacted for anonymity.} 
%We obtained authentic, human-authored CVs using Prolific, %a platform which enables stricter demographic and quality control in exchange for higher payment compared to other popular platforms.
On Prolific\footnote{\url{https://www.prolific.co/}}, we hired 2,000 participants (50\% women) who were (1)~US American and live and work in the US; and (2)~in full-time employment. After answering a number of screening questions, %, out of which we only use gender and the occupation they work in in this study. 
participants composed a CV ``pretending that you were applying for your next promotion''. We specifically asked our participants to copy from their existing CV, instead of write an entirely new CV to mimic real-world CVs as closely as possible. It was encouraged to anonymize information wherever possible, but otherwise craft a CV as realistic as possible given their current situation. 
% number of scredemographic questions We asked our participants to copy and paste contents from their already existing CV, whilst anonymising their names and the names of their affiliated organisations. 
For a uniformed structure, we segmented the CV submission into five parts, each as a free-text box: (1) an optional professional summary/career objective, (2) professional experience, (3) education; (4) skills and attributes; (5) optional certifications/qualifications.

\paragraph{Quality control and preprocessing}  We removed responses based on very short (long) response times and non-English text %that were not identified as not in the English language 
($\sim10\%$), retaining 1,789 CVs (50.5\% female). We tokenized, lemmatized and POS-tagged all text, removed stop words, and concatenated the five CV segments. We identified pronouns and named entities.\footnote{Including all entity types covered by SpaCy's default entity tagger.} All preprocessing was done using SpaCy's default English models.%\footnote{\url{https://spacy.io/}}

\paragraph{Data sharing} In line with our IRB approval, we release a deidentified version of our data set to individual researchers. Further details are in the Ethics Statement. Appendix~\ref{app:consent} contains the consent form.

\section{Gender-associated word choice}
\label{sec:tfidf_qual}
% Table~\ref{tab:tfidf_qual} summarizes our qualitative analysis of gender-specific language per occupation. We cover two female-dominated occupations (Education, Healthcare), two male-dominated (Computer/mathematical occupations and Management) and two gender-balanced (Business/Finance and Sales). We show the top 1\% TFIDF terms per gender that are {\it not} included in the equivalent list of the opposite gender. We provide the same analysis overall across all 16 occupations + `Other' (bottom of Table~\ref{tab:tfidf_qual}).
We qualitatively analyze gender-associated word choice in 6 (out of 16) occupations: 2$\times$ female dominated (Education, Healthcare); 2$\times$ male dominated (Computer/maths, Management) and 2$\times$ balanced (Business/finance, Sales). 

We first obtain the top 1\% of TFIDF-ranked unigrams for both men's ($M$) and women's ($F$) CVs. We then retain terms in these two sets unique to $M$ (and conversely, unique to $F$) as terms  highly associated with only one gender. Due to space constraints, we present the full results in Table~\ref{tab:tfidf_qual} in Appendix~\ref{app:tfidf}.

In men-dominated occupations, men-associated terms are `scientistic' (\texttt{engineer}, \texttt{developer}, \texttt{database}), or relate to leadership/tactics (\texttt{leadership},  \texttt{planning}); women-associated terms relate to interpersonal skills (\texttt{community}, \texttt{communication}, \texttt{social}). For women-dominated occupations, terms more likely to be used by women include those related to support and teamwork (\texttt{help}, \texttt{assistant}, \texttt{aid}); whereas men use terms which are again `scientistic' and exhibiting leadership (\texttt{physician}, \texttt{lead}, \texttt{manager}).

The overall, across-occupation, pattern is not dissimilar to the occupation-stratified analyses above. This is consistent with sociological studies which have shown that men are often assessed by their competence and leadership qualities, whereas women are often assessed by their 'likeability' (i.e., their personal characters) \cite{Eagly2002}. On the contrary, women who show ambition and competitiveness are often penalised for violating traditional feminine stereotypes \cite{phelan2008}. Such biased judgments are likely to discourage women to use words to describe their expertise and use more communal words instead. 

Note, however, that these differences arise not only from lexical choice, but also from real world differences {\it within} an occupational group, where the genders distribute differently across work tasks and finer-grained roles: women tend to have lower-ranking jobs, and specific occupations within the broader groups will exhibit different gender skews. Results for the Education occupation illustrate this well, where men-associated terms are dominated by technology and leadership (\texttt{microsoft}, \texttt{lead}, \texttt{technology}), while women-associated terms focus on early education and support (\texttt{child}, \texttt{elementary}, \texttt{social}). See underlined terms in Table~\ref{tab:tfidf_qual} for further examples.

\section{Power and Agency in CVs}
\label{ssec:powerandagency}
Do men and women differ in the way they present themselves in a CV? We compare the extent of {\it power} and {\it agency} implied in the verbs used by male and female applicants. %Focussing on verbs allows us to abstract away locations or institutions as well as occupation-specific concepts or technologyes, and instead focus on the self-presentation. 
We apply \citet{sap-etal-2017-connotation}'s connotation frames of power and agency, which associate verbs with the reactions they evoke the reader~\cite{rashkin-etal-2016-connotation}. By focussing on verbs, we abstract away from (named) entities with a strong occupation association and focus on self-presentation \cite{Goffman1959}. We consider all transitive verbs in CVs. Given the content (primarily focused on the author) and style (listings, incomplete sentences) of CVs, we assume that the agent of every verb is the author. The {\bf power} dimension distinguishes verbs where the agent (subject) has more ($A{>}T$; `lead'), less ($A{<}T$; `assist'), or equal ($A{=}T$; `care') power to the theme (object). The {\bf agency} dimension categorizes verbs as high ($+$; `support'), low ($-$; `wait') or neutral ($neu$; `access') agency. We use \citet{sap-etal-2017-connotation}'s frame-labeled data set of 2K English verbs. 48\% of verb types in our CVs are in the labeled data set (conversely, 57\% of frame-labeled verbs occur in our CVs). The numbers are comparable across genders.

\begin{figure}
    \centering
    \includegraphics[width=\columnwidth,clip,trim=0 0.4cm 0 0]{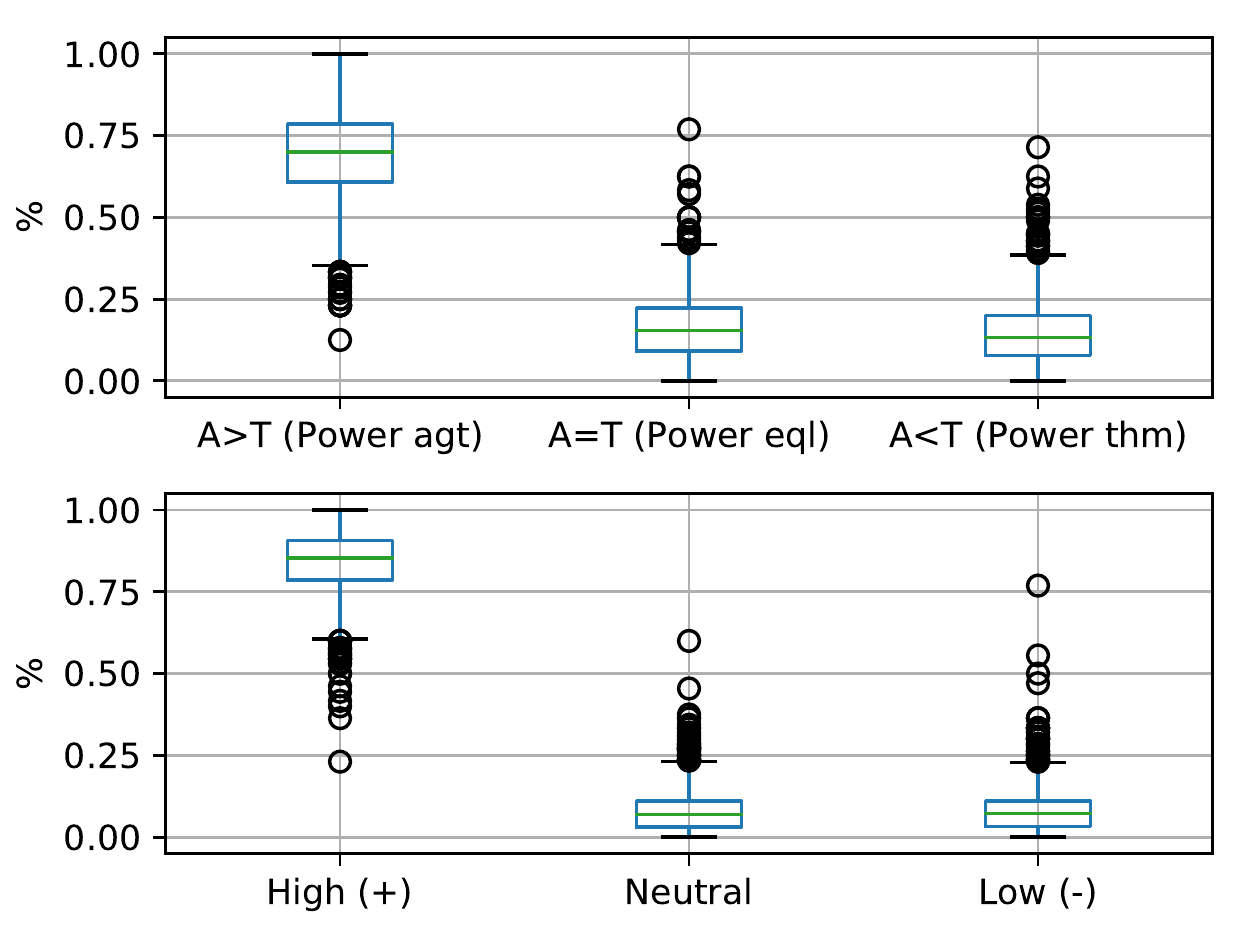}
    \caption{Distribution of power (top) and agency (bottom) verbs in our data. Each dot corresponds to one CV. }
    \label{fig:power_agency}
\end{figure}

\paragraph{Overall label distribution} We restrict our analysis to CVs with at least 10 and at most 100 verbs (N=1503, 53\% women) to reduce the impact of outliers,\footnote{Noting that the results hold with all data points included.} and retrieve the power and agency label of each verb that is included in the labeled set.  Figure~\ref{fig:power_agency} shows the overall distribution of power and agency levels in our data set. Consistent with prior work~\cite{sap-etal-2017-connotation}, and unsurprising given the data domain, we observe a dominance of agent-power and high agency verbs. 

\paragraph{Gender differences} We measure the statistical dependence of power/agency levels (independent variables) on the gender of the CV author (dependent variable) fitting a logistic regression.\footnote{\url{https://www.statsmodels.org/dev/generated/statsmodels.discrete.discrete_model.Logit.html}} Each CV is represented as a count vector over the 3 power and 3 agency categories, while controlling for CV length (number of words) and occupation (cf. Appendix~\ref{sec:app:occupations}). We standardized features for better interpretability of $\beta$, and coded Man as 0, and Woman as 1. Table~\ref{tab:significance} shows that women use equal power (A=T) and theme power (A<T) verbs significantly more often than men. Examples for equal power verbs that are more frequently used by females than males are \{\texttt{complete, perform, analyze, assess}\}, and for theme power verbs \{\texttt{assist, learn, need, serve%, attend
}\}. Gender difference for use of agent-power verbs is insignificant.

We also find that both high (+) and low \mbox{(-)} agency verbs are more frequently used by men.\footnote{Noting that the significance for the differences in agency do not hold after Holm–Bonferroni correction \cite{Holm1979-mg} for multiple comparisons.} Male-associated positive agency verbs include \{\texttt{employ, reduce, rate, acquire, exceed}\} while male-associated negative agency verbs include \{\texttt{address, expect, stay, relate}\}. There are no significant differences in the use of neutral agency verbs.

% \paragraph{Qualitative analysis} From the classes of verbs which are statistically-significant in Table \ref{tab:significance}, we notice significantly more "positive agency" and "negative agency" verbs used in male CVs. In particular, words such as \textit{monitor, exist, expect}, and \textit{stay}, which somewhat independently corroborate findings found in studies of gender bias in large language models TODO CITATION

\subsection{Discussion}
% \sheilla{or Marc: can you say this more nicely and add references? More?} Thanks Sheilla!
Women use more verbs in CVs that associate low power with the agent (CV author). Broadly, this agrees with prior work revealing that women are portrayed as less powerful in fiction movies~\cite{sap-etal-2017-connotation}. It also links into results from sociology revealing that leadership qualities (strength, assertiveness) are evaluated more positively in men, than in women~\cite{Eagly2002,Rudman2001}. Similarly, men are often rated more favorable than women given the same qualification, which might lead women to elaborate more on their education and qualifications \cite{Njoto2022-ISTAS-inpress}. Indeed, we find the Education section in female-authored CVs to be on average 15\% longer than in male (247 vs 214 words), the Qualification and Training section 24\% longer (144 vs 116 words), while the Professional Experience sections are of similar length (+2\%; 1109 vs 1034 words).

Notably, this may be because, on average, women in the US have attained a higher level of education that men~\cite{pew}.
% \footnote{https://www.pewresearch.org/fact-tank/2021/11/08/whats-behind-the-growing-gap-between-men-and-women-in-college-completion/)}. 
However, it also reflects that women tend to get more education for the same job, and tend to be overly qualified for similar positions. Women access more education but also need more education and training for the same job~\cite{campbell2022he}. 

\begin{table}[]
    \centering
    \begin{small}
    \begin{tabular}{lrlc}
      \toprule
      \multicolumn{1}{c}{Feature} & \multicolumn{1}{c}{$\beta$} & \multicolumn{1}{c}{p} \\\midrule
Power($A>T$) & 4.57 & 0.140\\
Power($A=T$) & 10.78 & 0.006 ** & {\bf F}\\
Power($A<T$) & 15.25 & 0.000 ** & {\bf F}\\\midrule
Agency($+$) & -6.35 & 0.032 * & {\bf M}\\
Agency($neu$) & 2.34 & 0.589\\
Agency($-$) & -11.72 & 0.007 * & {\bf M}\\\bottomrule
    \end{tabular}
    \end{small}
    \caption{Association of power/agency with binary CV author gender via coefficients ($\beta$) and significance estimates (p) of a logistic regression, after controlling for CV length and occupation. The final column indicates direction of association (male=0, female=1). * indicates statistical significance at $p<0.05$, while ** additionally confirms significance after Holm–Bonferroni  correction for multiple comparisons.}
    \label{tab:significance}
\end{table}

\section{Gender prediction from CV text}
\label{sec:prediction}
Sections~\ref{sec:tfidf_qual} and~\ref{ssec:powerandagency} explored gender-speciric content framing differences CVs which may impact judgment the reader (or hirer). We next quantify the susceptibility of representative predictive models to gender information in CVs. We use the task of binary author gender prediction based on the text of the CV as a diagnostic tool to assess the extent to which models can infer gender information from CVs. We explicitly caution against using this task as a ML benchmark (cf., Ethics Statement).

% Having shown that differences in word choice and framing exist between male and female-authored CVs, and acknowledging that we next quantify the susceptibility of predictive models to gender information. 

We test the following binary classifiers: a linear SVC with L1 regularization, which by design learns sparse and interpretable features; % allowing to inspect the most predictive features (words) for each class; (2)~
a logistic regression classifier (LR); and a fine-tuned RobERTa-based classifier built on pre-trained RoBERTa uncased~\citep{https://doi.org/10.48550/arxiv.1907.11692}, fine-tuned for two epochs with a learning rate of \num{4e-5}.\footnote{BERT uncased performed slightly below RoBERTa.} We use TFIDF features ($|X|$=5000) for LR and SVM and plain text for RoBERTa.  We include a random uniform, and a majority baseline and run all models with 5-fold cross-validation.
% as well as a more powerful non-linear BERT-based classifier. We trained binary classifiers based on BERT\citep{47751} and RoBERTa\citep{https://doi.org/10.48550/arxiv.1907.11692} pretrianed models. Both of them are uncased, the learning\_rate is \num{4e-5} and every training contains 2 epochs.

We test our models on three versions of our CVs.\footnote{Like in Section~\ref{ssec:powerandagency} we remove CVs with fewer than 10 or more than 100 verbs for consistency.} All versions are lemmatized and stopwords were removed. (1) the full data set with all lemmas from all CV sections ({\bf Full}); (2) mask names and pronouns to remove explicit gender indicators ({\bf -PER}); (3) remove {\it all} named entities ({\bf -NE}), to abstract away from institutional information such as single-gender schools which may carry implicit information about the gender of the applicant.

Training these models on the full data sets (D-Full, N=1503) set will inevitably add a confounding factor of occupation-specific terms: most occupations are substantially gender-skewed in their workforce. To remove this confound, we create a version of each data set with a gender-balanced  set of CVs {\it for each occupation} (D-Balanced, N=1118). We report results as macro averaged F1 scores, as presented in Table~\ref{tab:gender_prediction}. 
\begin{table}[]
    \centering
    %     \toprule
    %   & \multicolumn{2}{c}{Macro F1} \\
    %     Model & Full & Balanced\\\midrule
    %     Random U& 0.50  ($\pm$0.0)& 0.48($\pm$0.0)\\
    %     Majority & 0.34 ($\pm$0.0)& 0.33  ($\pm$0.0)\\\midrule
    %     SVC Full & 0.69 ($\pm$ 0.001)& 0.64 ($\pm$0.1)\\
    %     SVC -PER& 0.69 ($\pm$ 0.000)& 0.61 ($\pm$0.0)\\ 
    %     SVC -NE & 0.67 ($\pm$ 0.002)& 0.59 ($\pm$0.002)\\\midrule
    %     LR Full & 0.72 ($\pm$ 0.001)& 0.66 ($\pm$0.0)\\
    %     LR -PER & 0.73 ($\pm$ 0.001)& 0.66 ($\pm$0.0)\\
    %     LR -NE & 0.71 ($\pm$ 0.002)& 0.63 ($\pm$0.0)\\\midrule
    %     BERT Full & 0.74 ($\pm$ 0.0)& 0.67 ($\pm$0.001)\\
    %     BERT -PER & 0.73 ($\pm$ 0.0)& 0.57 ($\pm$0.004)\\
    %     BERT -NE & 0.73 ($\pm$ 0.0))& 0.62 ($\pm$0.001)\\\midrule
    %     RoBERTa Full & 0.75 ($\pm$ 0.0)& 0.71 ($\pm$0.001)\\
    %     RoBERTa -PER & 0.75 ($\pm$ 0.0)& 0.57 ($\pm$0.042)\\
    %     RoBERTa -NE & 0.73 ($\pm$ 0.0)& 0.66 ($\pm$0.0)\\
    %     \bottomrule
\begin{small}
    % \begin{tabular}{l|ll}
    % \toprule
    % %   & \multicolumn{2}{c}{Macro F1} \\
    %     Model & D-Full & D-Balanced\\\midrule
    %     Random U& 0.50  ($\pm$0.0)& 0.48($\pm$0.0)\\
    %     Majority & 0.34 ($\pm$0.0)& 0.33  ($\pm$0.0)\\\midrule
    %     SVC Full & 0.69 ($\pm$ 0.001)& 0.64 ($\pm$0.1)\\
    %     LR Full & 0.72 ($\pm$ 0.001)& 0.66 ($\pm$0.0)\\
    %     % BERT Full & 0.74 ($\pm$ 0.0)& 0.67 ($\pm$0.001)\\
    %     RoBERTa Full & 0.75 ($\pm$ 0.0)& 0.71 ($\pm$0.001)\\\midrule
    %     SVC -PER& 0.69 ($\pm$ 0.000)& 0.61 ($\pm$0.0)\\ 
    %     LR -PER & 0.73 ($\pm$ 0.001)& 0.66 ($\pm$0.0)\\
    %     % BERT -PER & 0.73 ($\pm$ 0.0)& 0.57 ($\pm$0.004)\\
    %     RoBERTa -PER & 0.75 ($\pm$ 0.0)& 0.57 ($\pm$0.042)\\\midrule
    %     SVC -NE & 0.67 ($\pm$ 0.002)& 0.62 ($\pm$0.00)\\
    %     LR -NE & 0.71 ($\pm$ 0.002)& 0.66 ($\pm$0.0)\\
    %     % BERT -NE & 0.73 ($\pm$ 0.0))& 0.62 ($\pm$0.001)\\
    %     RoBERTa -NE & 0.73 ($\pm$ 0.0)& 0.66 ($\pm$0.0)\\
    %     \bottomrule
    % \end{tabular}
    \begin{tabular}{lll}
    \toprule
    %   & \multicolumn{2}{c}{Macro F1} \\
        \multicolumn{1}{c}{\bf Model} & \multicolumn{1}{c}{\bf D-Full} & \multicolumn{1}{c}{\bf D-Balanced}\\\midrule
        Random U& 0.50  ($\pm$0.00)& 0.48($\pm$0.00)\\
        Majority & 0.34 ($\pm$0.00)& 0.33  ($\pm$0.00)\\\midrule
        SVC Full & 0.69 ($\pm$ 0.03)& 0.64 ($\pm$0.03)\\
        LR Full & 0.72 ($\pm$ 0.03)& 0.66 ($\pm$0.02)\\
        % BERT Full & 0.74 ($\pm$ 0.0)& 0.67 ($\pm$0.001)\\
        RoBERTa Full & 0.75 ($\pm$ 0.02)& 0.71 ($\pm$0.03)\\\midrule
        SVC -PER& 0.69 ($\pm$ 0.01)& 0.61 ($\pm$0.03)\\ 
        LR -PER & 0.73 ($\pm$ 0.03)& 0.66 ($\pm$0.02)\\
        % BERT -PER & 0.73 ($\pm$ 0.0)& 0.57 ($\pm$0.004)\\
        RoBERTa -PER & 0.75 ($\pm$ 0.01)& 0.57 ($\pm$0.20)\\\midrule
        SVC -NE & 0.67 ($\pm$ 0.02)& 0.62 ($\pm$0.01)\\
        LR -NE & 0.71 ($\pm$ 0.02)& 0.66 ($\pm$0.02)\\
        % BERT -NE & 0.73 ($\pm$ 0.0))& 0.62 ($\pm$0.001)\\
        RoBERTa -NE & 0.73 ($\pm$ 0.02)& 0.66 ($\pm$0.01)\\
        \bottomrule
    \end{tabular}
    \end{small}
    \caption{Predicting the gender (M,F) of an author of a CV. Macro-averaged F1 score ($\pm$standard deviation) from 5-fold cross-validation.}
    % -PER: names and pronouns removed. -NE: all named entities removed. D-Full: trained on the full data; D-Balanced: trained on a data subset which is gender-balanced per occupation.}
    \label{tab:gender_prediction}
\end{table}

\paragraph{Results} We test whether gendered information is encoded in classifiers trained on data with varying amounts of gender-indicative information.
% -- by excluding all entities and by removing the occupation confounding factor by gender-balancing the data. 
A perfectly gender-agnostic model would perform en par with the baselines.

Table~\ref{tab:gender_prediction} shows that all classifiers outperform the baselines substantially, both when trained on D-Full as well as on D-Balanced, where the occupation proxy gender is reduced. In line with prior work~\cite{de2019bias}, we find that `scrubbing' names and pronouns as explicit gender indicators (-PER) has negligible impact.  Removing all named entities reduces classifier performance, but it remains well above random. RoBERTa outperforms the linear models in the Full data condition (left column), but shows unstable performance in the Balanced condition (right column) presumably due to the smaller data set leading to overfitting (note the high std in the -PER condition).  Overall, the findings highlight the importance of considering gendered language signals beyond explicit indicators, i.e.,~that simple methods like removal of names~\cite{Manikandan} does not imply absence of gender information. Table~\ref{tab:svcfeatures} (Appendix~\ref{app:features}) lists the 20 most predictive features  for the linear SVC trained on the D-Balanced, when trained on the full data (top) and the entity-redacted data (bottom). The features from the full data include entities like state names (\texttt{Indianapolis, Colorado}). Even after gender balancing per occupation, stereotypically associated features with women (\texttt{child}, and ``soft" attributes like \texttt{attitude, assist, document}) and men (\texttt{supervise, technology}) emerge.

% Unsurprisingly, names are among the most predictive features in the full data.\footnote{the concrete names John and Jane presumably a result of annotators anonymizing their CVs, as encouraged.} The remaining features reflect both occupational gender imbalance 
% (machine, technology for males and child, service for females) as well as actions and characteristics typically associated with the genders (executive, leadership for males; communicating, creating, answering/assisting for females), which was hitherto alluded to in Section \ref{sec:powerandagency}. The predictive features based on the  gender-balanced CV data set are dominated by verbs, and female features (unlike male) include several terms relating to education (honors, resume, training, ...). 

\section{Discussion}
\label{sec:discussion}
We presented a data set of 1.8K authentic, US-English CVs across 16 occupations, aligned with self-reported binary gender of the author. This data set allowed us to inspect features of men- and women-specific language in CVs, while controlling for the confounding factor of occupation: most occupations are heavily gender-skewed.

This paper connects the concept of framing, i.e.,~influencing readers of a document through careful choice of words~\cite{entman2007framing}, with existing power discrepancies between men and women in western society in general, and the job market specifically~\cite{rudman1998self}. We showed that women use verbs that imply lower power significantly more often than men, even after controlling for  occupation. Subtle changes in word choice have been shown to impact human perception, reaction and choice~\cite{kahneman2013choices}. In the context of {\it human} hiring, this suggests that (a)~removing explicit gender indicators is insufficient; and (b)~further support for sensitizing both hirers and applicants to subconscious bias.

We further trained classifiers to predict binary author gender based on CV text, in scenarios where gender proxy information was removed by gender-balancing the training data and/or removing named entities. We show that classifiers perform significantly above chance across all settings, confirming that subtle gender signal remains. This result is expected, and in line with prior research~\cite{de2019bias}, but for the first time shown directly on data more akin to application materials presented to human and automatic hirers.

Our experiments retain a confounding factor of job type within an occupational group: within an occupation, women tend to have lower-ranking jobs; and within our 16 broad occupational groups, different specific occupations will exhibit different gender skews. In Section~\ref{sec:tfidf_qual}, we inspected gender-associated word choice in 6 most frequent occupations in our data set, finding that across occupations, `scientistic' terms (\texttt{engineer}, \texttt{developer}, \texttt{database}) and leadership terms (\texttt{leadership}, \texttt{administration}, \texttt{planning}) are more associated with male CVs; while women are more likely to mention interpersonal skills, support, or teamwork  (\texttt{community}, \texttt{communication}, \texttt{social},\texttt{help}, \texttt{assistant}, \texttt{aid}), typically associated with administrative roles. Consequently, gender signals in CVs not only originate from lexical choice, but also also reflect real-world differences in work tasks and position levels. Disentangling these factors is an important direction for future work.

In sum, we maintain that perpetuated gendered patterns embedded in CVs can bias both human and automated hiring, and that the naive use of ML methods bears the risk of exacerbating bias: by picking up spurious associations on different levels from explicit gender information (names, hobbies) to subtler word choice (the level `power' or `agency').
% : to wit, via whatever statistical pattern (of `powerful' or `agentic' language) it is trained on.
Suggestions for further work include usability studies and social-psychological interventions for users of recruitment software, for \textit{both} job applicants and decision makers. Interventions could include `nudges' in the user experience flow, informing users about potential gender signals being encoded in their data, and suggestions of strategies to mitigate or minimise this. As our findings suggest, scrubbing names and entities off the CVs is not effective in de-gendering CVs for fairer recruitment decisions, and should not be used as the be-all-and-end-all in promoting fair hiring, as often is the wont of current initiatives.

\section*{Ethics Statement}
%\todomc{Marc!}
This study was approved by the University of Melbourne ethics board (Human Ethics Committee LNR 3A), Reference Number 2022-22062-32741-5, and data acquisition and analysis has been taken out to the according ethical standards. Our data was collected via Prolific. The crowdworkers (annotators) in the project were paid £3.75 for a median of 11 minutes of work, which is slightly above minimum hourly wage and reflects adequate compensation for the time spent. Appendix~\ref{app:consent} contains the consent form presented to annotators before the task. Prolific allows us to record information anonymously without personally identifiable data. As part of CV generation, our crowdworkers were instructed to exclude their names and the names of their affiliated organisations from their drafted CVs.

To enable future research in this area, we plan to release an anonymized and deidentified version of our data to individual researchers where names, emails, addresses, phone numbers and all named entities are redacted (the [-NE] version used in this paper). The data will contain the redacted CV text and self-identified gender label only. Interested researchers will sign an agreement form stating that they (1)~will not share the data with anyone else; (2)~will delete the data upon completion of the research or after 1 year whichever comes first. 

This paper investigated the language differences between men and women authored CVs. Gender information was identified by the CV authors and no gender-inference was applied anywhere in the paper. We acknowledge that a binary notion of gender is not representative of the concept. In addition, we acknowledge that our study excludes a large portion of the population which does not identify to a cis-normative group. We emphasized throughout the paper that our findings hold for self-identifying men and women only, and that our methodology in principle extends to a more inclusive set of gender groups, conditioned on the availability of reliable data.

We used the task of gender prediction from CV data as a benchmark to assess the amount of gendered information retained in ML models after various strategies to remove gender proxy information. We do not endorse this task in general, and accordingly do not release pre-trained models to the public.

% \textcolor{red}{TODO: CAUTION AGAINST GENDER PREDICTION -- WE RELEASE CODE / MODELS ONLY TO DIRECT CONTACTS.}

% Care was taken to ensure that ethical standards have been met in the process of data acquisition and analysis, in accordance with [ethics board name redacted for review].

%  The data will only contain the deidentified CV text and associated gender label. Researchers who want to access the data will sign an agreement to not share the data further, and delete it after the research is completed or after one year (whichever comes first).
%Annotator treatment, privacy, copyright, \dots. I suggest to share an ANONYMIZED version of the data upon request (does ethics allow this)?

\section*{Acknowledgments}
We thank the reviewers for their very insightful feedback. This work was partially funded by the Seed Funding scheme of the Melbourne Center for Data Science.

\bibliography{emnlp2022}
\bibliographystyle{acl_natbib}

\appendix

\section{Prolific consent form}
\label{app:consent}
\texttt{
    Consent Form}

\texttt{
I consent to participate in this project, the details of which have been explained to me.}

\texttt{
I understand that the purpose of this research is to investigate how to produce a CV that increases success in recruitment.}

\texttt{
I understand that my participation in this project is for research purposes only. }

\texttt{
I acknowledge that the possible effects of participating in this research project have been explained to my satisfaction.}

\texttt{
In this project, I will be required to draft a CV to apply for a promotion.}

\texttt{
I understand that my participation is voluntary and that I am free to withdraw from this project anytime without explanation or prejudice and to withdraw any unprocessed data that I have provided.}

\texttt{
I understand that the data from this research will be stored at the University of Melbourne and will be destroyed after 5 years. }

\texttt{
I have been informed that the confidentiality of the information I provide will be safeguarded and subject to any legal requirements; my data will be password protected and accessible only by the university-approved researchers.}

\texttt{
I understand that all data recorded is anonymized.}

\texttt{
I understand that after I sign and return this consent form, it will be retained by the researcher.}

\texttt{
By clicking the checkbox below, you are signing the Consent Form.
}

\section{Occupations}
\label{sec:app:occupations}
The occupational code was taken from the General Social Survey,\footnote{https://gss.norc.org/} as the Standard Occupational Classification System widely used in English-speaking official surveys. categorization  Table~\ref{tab:occupations} lists all occupations in our data set, together with the total number of CVs and gender ratio. "Other" occupations arise from free-text entries. Some examples include dog caretaker, musician, transport and logistics manager, pilot, automotive designer, among others. For reference, we also include gender rations from the 2021 US Labor Statistics in the rightmost column of Table~\ref{tab:occupations}. In general, the gender skew in our data set agrees with the US statistics, with some deviations expected given our limited sample.

\begin{table*}[]
    \centering
    \begin{tabular}{p{8cm}cccc|c}
    \toprule
occupation & Man & Woman & \% F & Total & \% F USLS \\\midrule
{\bf Business and financial operations occupations} & 117 & 85 & { 0.42} & 202 & 0.55\\
{\bf Computer and mathematical occupations} & 152 & 49 & { 0.24} & 201 & 0.26\\
{\bf Educational instruction and library occupations} & 64 & 117 & { 0.65} & 181 & 0.74\\
{\bf Healthcare support occupations} & 58 & 118 & 0.67 & 176& 0.85\\
{\bf Management occupations} & 101 & 58 & 0.36 & 159& 0.52\\
{\bf Sales and related occupations} & 65 & 70 & 0.52 & 135& 0.62\\
Arts, design, entertainment, sports, and media occupations & 58 & 64 & 0.52 & 122& 0.50\\
Office and administrative support occupations & 43 & 78 & 0.64 & 121& 0.72\\
Other & 61 & 53 & 0.46 & 114& --\\
Life, physical, and social science occupations & 33 & 62 & 0.65 & 95& 0.61\\
Architecture and engineering occupations & 56 & 18 & 0.24 & 74& 0.24\\
Community and social service occupations & 15 & 44 & 0.75 & 59& 0.72\\
Food preparation and serving related occupations & 20 & 33 & 0.62 & 53& 0.59\\
Legal occupations & 28 & 22 & 0.44 & 50& 0.42\\
Personal care and service occupations & 8 & 21 & 0.72 & 29& 0.72\\
Protective service occupations & 5 & 5 & 0.5 & 10& 0.50\\
Farming, fishing, and forestry occupations & 2 & 6 & 0.75 & 8& 0.75\\\midrule
Total & 886 & 903 & 0.5 & 1789 & 0.53\\
\bottomrule
\end{tabular}
    \caption{Occupations with number of CVs and proportion of female participants (\%F) in our CV data set. The occupations included in our analyses in Appendix~\ref{app:features} are bold-faced. \% F USLS are the official percentages of female employees per occupation taken from the US Labor Statistics (2021).}
    \label{tab:occupations}
\end{table*}

%%%%%%%%%%%%%%%%

\begin{table*}[]
    \centering
    \begin{small}

    \begin{tabular}{p{2.3cm}p{3.2cm}p{9cm}}
    \toprule
    {\bf Gender balance} & {\bf Occupation} & {\bf Terms}\\\midrule

  {\bf Gender-balanced} & Business and financial operations occupations & 
  $M\setminus F$: insurance, sale, analysis, computer, word, lead, knowledge, master, data, product, time, operation, tax\\%\cline{3-3}
  & & 
  $F\setminus M$: student, august, accountant, art, high, information, public, use, training, software, development, research, program\\\midrule 
  & Sales and related occupations & 
  $M\setminus F$: major, proficient, problem, engineering, june, computer, account, issue, lead, use, goal, responsibility, technical, develop, representative, operation, support, strategy\\%\cline{3-3}
  & & $F\setminus M$: position, degree, industry, august, english, excel, excellent, leadership, strong, student, art, medium, create, study, maintain, good, employee, communication\\\midrule 
  
  {\bf Men-dominated} & Computer and mathematical occupations & 
  $M\setminus F$: gpa, application, database, office, window, microsoft, high, server, issue, engineer, web, security, include, developer, product, develop, network, college\\%\cline{3-3}
  & & 
  $F\setminus M$: student, art, analysis, lead, java, ms, python, course, sale, master, social, communication, spanish, css, time, html, graduate, ability, maintain, research\\\midrule 
  & Management occupations & 
  $M\setminus F$: california, excel, leadership, engineering, ability, computer, company, information, planning, marketing, software, technology, product, time, administration, degree, support, design\\%\cline{3-3}
  & & $F\setminus M$: staff, position, member, community, diploma, medium, account, research, health, hi, job, master, study, public, social, create, graduate, proficient\\\midrule 
   
  {\bf Women-dominated} & Educational instruction and library occupations & 
  $M\setminus F$: project, level, office, datum, \underline{microsoft}, \underline{computer}, \underline{lead}, new, \underline{software}, \underline{technology}, lesson, business, \underline{develop}, class, proficient\\%\cline{3-3}
  & & 
  $F\setminus M$: gpa, degree, library, member, \underline{child}, community, award, create, \underline{elementary}, honor, psychology, development, \underline{social}, \underline{help}, write\\\midrule 
  & Healthcare support occupations & 
  $M\setminus F$: physician, various, equipment, project, member, volunteer, community, emergency, information, manage, manager, 
department, able, ensure, american\\%\cline{3-3}
  & & $F\setminus M$: gpa, august, assistant, aid, problem, assist, june, art, microsoft, customer, paste, time, cpr, therapy, role\\\midrule \midrule
\multicolumn{2}{l}{\textbf{Overall: across all occupations}} & $M\setminus F$:
web, server, improve, production, good, day, engineering, tool, build, policy, degree, check, increase, test, 
technology, meet, senior, master, material, security, engineer, control, procedure, sql, solution, point, 
administration, performance, quality, database, network, equipment, data, strategy, user, testing \\
& & $F\setminus M$:
strong, psychology, able, conduct, word, care, position, organize, prepare, art, document, intern, national, 
coordinate, resource, online, record, schedule, teacher, space, course, gpa, teach, english, child, current, event,
class, store, volunteer, september, meeting, honor, individual, content, study, phone \\\bottomrule
    \end{tabular}
    \end{small}
% \vspace{10pt}
% \vspace{10pt}

% \begin{small}
% \begin{tabular}{p{5.5cm}p{9cm}}
% \toprule
%  &  Terms\\\midrule
% \textbf{Overall: across all occupations} & $M\setminus F$:
% web, server, improve, production, good, day, engineering, tool, build, policy, degree, check, increase, test, 
% technology, meet, senior, master, material, security, engineer, control, procedure, sql, solution, point, 
% administration, performance, quality, database, network, equipment, data, strategy, user, testing \\\\

%  & $F\setminus M$:
% strong, psychology, able, conduct, word, care, position, organize, prepare, art, document, intern, national, 
% coordinate, resource, online, record, schedule, teacher, space, course, gpa, teach, english, child, current, event,
% class, store, volunteer, september, meeting, honor, individual, content, study, phone \\\bottomrule
% \end{tabular}
% \end{small}
    \caption{Qualitative analyses of mutually-exclusive terms within the top 1\% of unigrams, by tf-idf ranking:     (top) stratified across occupations; and (bottom) across all occupations. $M\setminus F$ denotes the set of terms in the top 1\% of male CV unigrams, which are not in the corresponding top 1\% of female CV unigrams.
    Conversely, $F\setminus M$ denotes the set of terms found in the top 1\% of female CV unigrams, which are not in the top 1\% of male CV unigrams. The underlined examples are discussed in Section~\ref{sec:tfidf_qual}.}
    % full CVs (top two rows) and named entity-redacted CVs (bottom two rows).}
    \label{tab:tfidf_qual}
\end{table*}

\section{Gender-associated word choice}
\label{app:tfidf}
Table~\ref{tab:tfidf_qual} shows the full results of gender-specific top 1\% TFIDF terms per occupation (top), and overall across our whole CV data set (bottom).

\section{Classifier features}
\label{app:features}
\begin{table}
    \centering
    \begin{small}
    \begin{tabular}{p{\columnwidth}}
    \toprule
    % FULL\\\midrule
    %     %  {\bf male}: 
    %     %  john, volume, machine, organic, leader, journal, mar, improve, address, level, executive, parking, commercial, active, waste, utilize, technology, construction, missouri, instructor\\\midrule
    %     %  {\bf female}: answer, jane, social, donor, create, serve, microstrategy, solving, conduct, payable, child, file, guide, communication, ms, ny, operational, guest, know, research\\\midrule
    %      {\bf male -NE}: machine, technology, executive, year, leader, volume, improve, commercial, parking, engineer, digital, increase, waste, instructor, tool, responsibility, application, operating, observe, issue\\\midrule
    %      {\bf female -NE}: answer, donor, social, child, honors, serve, create, present, age, development, document, guest, communication, know, woman, club, research, correspondence, reference, role\\\midrule
    % BALANCED\\\midrule
   {\bf male}: level, indianapolis, instruct, reduce, th, clinical, troubleshoot, large, part, culinary, engineer, supervise, repair, improvement, shipping, technology, multiple, tool, business, regulatory\\
   {\bf female}: answer, attitude, reference, file, document, create, assist, role, colorado, children, child, coordinator, know, gain, interview, receivable, honors, woman, media\\\midrule
  {\bf male -NE}: machine, instruct, profit, clinical, also, supervise, observe, part, technology, leader, reduce, reduction, basic, instructor, opportunity, hold, review, people, regard, electrical\\
    {\bf female -NE}: check, attitude, document, core, medium, media, woman, answer, content, present, child, create, speaking, claim, resource, file, assist, resume, plan\\\bottomrule
    \end{tabular}
    \end{small}
    \caption{Most predictive features as learned by the binary SVC for gender prediction when trained on the Balanced data set full (top) or with NEs redacted (bottom).}
    % full CVs (top two rows) and named entity-redacted CVs (bottom two rows).}
        \label{tab:svcfeatures}
\end{table}

Table~\ref{tab:svcfeatures} lists the most predictive features for men (top) and women (bottom) as learnt by the linear SVC with L1 regularization when applied to the occupation-wise gender-balanced CV data set.

\end{document}